\documentclass[letterpaper, 10 pt, journal, twoside]{IEEEtran}

\usepackage{amsmath,amssymb,amsfonts}
\usepackage{gensymb}
\usepackage{algorithmic}
\usepackage{graphicx}
\usepackage{textcomp}
\usepackage[dvipsnames]{xcolor}
\def\BibTeX{{\rm B\kern-.05em{\sc i\kern-.025em b}\kern-.08em
    T\kern-.1667em\lower.7ex\hbox{E}\kern-.125emX}}

\usepackage{subcaption}
\usepackage[utf8]{inputenc}
\usepackage[T1]{fontenc}
\usepackage{url}

\usepackage{wasysym}

\usepackage{easyReview}
\usepackage[normalem]{ulem}
\usepackage{cancel}

\usepackage{makecell,tabularx,booktabs}
\newcolumntype{C}{>{\centering\arraybackslash}X} 
\newcolumntype{L}{>{\raggedright\arraybackslash}X} 
\newcolumntype{R}{>{\raggedleft\arraybackslash}X} 

\usepackage[style=ieee,backend=biber]{biblatex}

\usepackage{mathtools,amssymb,lipsum, nccmath}

\usepackage{xcolor}

\usepackage{booktabs}
\usepackage{array}
\usepackage{threeparttable}
\usepackage{tabularx}

\newif\ifshowedits
\showeditsfalse 

\newcommand{\removed}[1]{%
  \ifshowedits
    \textcolor{red}{\sout{#1}}%
  \else
    \ignorespaces
  \fi
}

\newcommand{\newtext}[1]{%
  \ifshowedits
    \textcolor{blue}{#1}%
  \else
    #1%
  \fi
}

\usepackage{colortbl}

\newenvironment{newtextblock}{%
  \begingroup
  \ifshowedits \color{blue} \arrayrulecolor{blue}\fi
}{%
  \arrayrulecolor{black}\endgroup
}

\begin{document}

\title{Hall Effect-Based Tactile Force Detection Sensor for Robot-Assisted Minimally Invasive Surgery \\}

\author{Charles DeLorey*~\IEEEmembership{Graduate Student Member,~IEEE,} Michael Brockdorff, Adam Metcalf~\IEEEmembership{Member,~IEEE}, Benjamin Calm\'e~\IEEEmembership{Member,~IEEE}, Pietro Valdastri~\IEEEmembership{Fellow,~IEEE,} and Dominic Jones*~\IEEEmembership{Member,~IEEE}
\thanks{This work was supported by the Engineering and Physical Sciences Research Council (EPSRC) under Grant UKRI180 (TARAS), and grant EP/Y010523/1 (UK-RAS Network Project: “OphthaTact")".} %
\thanks{The authors are with the STORM Lab, School of Electronic and Electrical Engineering, University of Leeds,  Leeds, UK. Email: \{\texttt{elchd, d.p.jones\}@leeds.ac.uk}}
}

\maketitle

\begin{abstract}
Robotic minimally invasive surgery has revolutionized surgical practice. However, as the surgeons are mechanically separated from the surgical tools, loss of \removed{force}\newtext{tactile} feedback occurs. We present a compact force sensor with three, millimeter scale sensing elements (taxels) that can be affixed to the grasping face of a robotic surgical tool. The sensor features a miniaturized design using Hall effect sensors and magnets embedded in a deformable elastomer contact layer. Sensor calibration is achieved with a 6 degree of freedom (DoF) parallel robot. Each taxel can detect applied normal and shear forces with average RMSE of \removed{1.13}\newtext{2.133} kPa. The calibrated sensor was validated on phantom and ex vivo porcine tissue\removed{with an RMSE of 2.39~kPa}, demonstrating \removed{force}\newtext{tactile} sensing during \removed{tissue} retraction as well as \removed{tissue} slip. This work introduces a small-scale Hall effect based tactile sensor, enabling sensing for force-sensitive robotic surgical applications.
\end{abstract}

\begin{IEEEkeywords}
Surgical robotics, force sensing, tactile sensing, grasping, perception for grasping and manipulation, soft sensors and actuators.
\end{IEEEkeywords}

\section{Introduction} 

Robot-Assisted Minimally Invasive Surgery (RAMIS) offers the gold-standard in surgical treatment for a range of procedures and provides improved visualization and dexterity with teleoperated instruments.
In traditional open surgery and manual laparoscopy, surgeons heavily rely on tactile sensations through their fingertips to assist in safe manipulation, feeling the tissue contact either directly or through the tool linkage \cite{tholey2005force}. 
Modern RAMIS systems, such as Intuitive Surgical's \textit{da Vinci} platform, physically decouple the surgeon from instrument–tissue interaction via remote teleoperation. 
This improves precision but removes natural tactile perception at the end-effector, leaving surgeons to infer the forces visually \cite{patel2022haptic}. 
Trauma from incorrect force application comes through two routes: excessive grasping can crush tissue leading to ischemia, \removed{(lack of blood flow)} while insufficient force risks slip and abrasion.
\removed{Compounding this problem, grip forces have been shown to vary between different instruments \cite{mucksavage2011differences}.}\newtext{Additionally, grip force in RAMIS has been shown to vary significantly between instrument as well as platform versions \cite{mucksavage2011differences}, with different tissue types requiring different target levels of force.}

Recently, the \textit{da Vinci} 5 (Intuitive Surgical) integrated on-instrument force sensing and visuohaptic feedback \cite{intuitive2024davinci5}. However, these capabilities only offer a \newtext{bulk kinesthetic} \removed{generalized} force feedback, with no information as to the contact forces the tissue experiences. Despite extensive research in surgical force sensing \cite{Schmidgall2025WillSurgery}, reliable local force measurement directly at the grasper–tissue interface remains unresolved. Existing approaches either estimate forces indirectly, or provide bulk interaction forces but are unable to sense compressive force on the tissue.

Force sensing can be introduced at several points along the RAMIS instrument chain: 1. motor-torque estimation at proximal actuators (indirect, model-dependent), 2. bulk sensing in the shaft (global reactions; e.g., da~Vinci~5) \cite{servais2025novel}, 3. sensors at \removed{or near} the trocar (extra-corporeal, decoupled from local contact), 4. wrist-mounted sensors (requiring kinematic back-projection), and 5. the contact face of the jaws.
\removed{Among these approaches}\newtext{Of these, on-jaw sensing} \removed{at the jaw contact face} provides the most direct access to \removed{reliable} local contact mechanics.  
Such on-jaw sensing is especially critical for observing contact pressures, distributed shear, and tissue-dependent friction behavior in highly variable surgical conditions.

\begin{figure}
    \centering
    \includegraphics[width=0.95\linewidth]{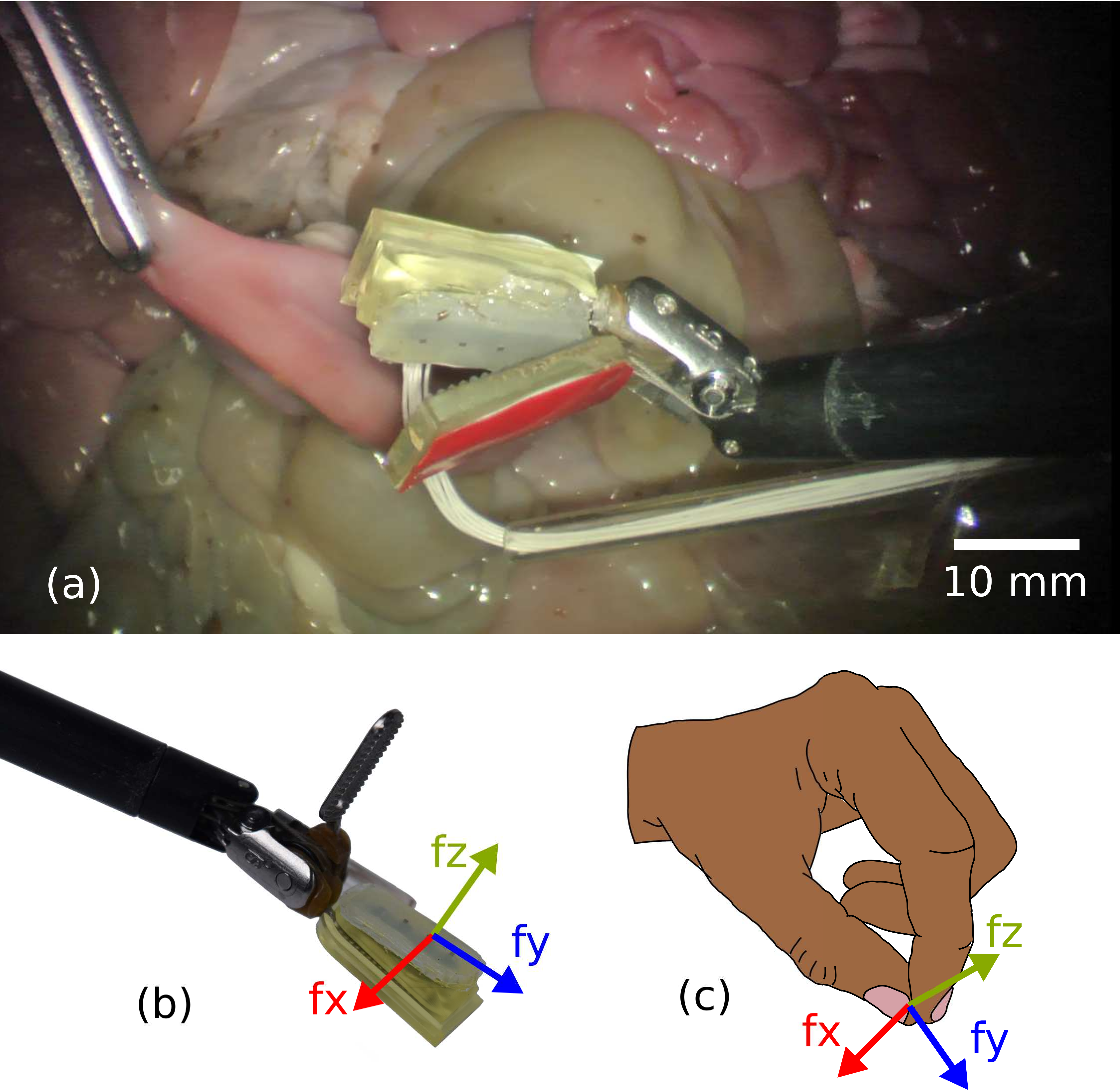}
    \caption{Force sensor in (a) freeform tissue grasping, (b) the assembled sensor on instrument and (c) analogous bioinspired fingertip pinch inspiration.}
    \label{fig:sensor-overview}
\end{figure}

Prior work has explored several transduction mechanisms for instrumented surgical graspers. 
Piezoresistive MEMS \removed{have been integrated into forceps with native tri-axial readout and compact die size, offering}\newtext{offer} high bandwidth at the cost of drift and the need for careful compensation and encapsulation \cite{hou2024piezoresistive}. 
Capacitive tactile films provide thin, conformable coverage with normal and shear sensing and low power, but are susceptible to fluids and humidity that shift the dielectric and demand shielding and packaging attention \cite{usman2025_sensorized_grasper, zhang2024tacscope}. 
Optical waveguide or imaging-based tactile devices sized for MIS tips achieve high spatial resolution and multi-axis estimation with remote optoelectronics, though maintaining optical paths and routing within tight curved jaws raises integration complexity \cite{li2024opticalwaveguide}. 
Fiber Bragg grating sensing enables electromagnetic interference-immune, multiplexed measurements for flexible endoscopic tools, yet requires extremely expensive interrogators and bonding strategies that can place sensing away from the true contact interface \cite{wang2024_fbg_measurement}. 
Finally, magnetic Hall-effect tactile approaches have demonstrated grasp-aware control with compact packages, electrical simplicity, \newtext{and} high sampling rates, but require compensation to mitigate stray-field and orientation biases \cite{waters2021utilising, waters2022incipient, du2023design}. \newtext{A comparison of these related works is shown in Table \ref{tab:sota-sensors}.}

\begin{table*}[!t]
\begin{newtextblock}
\renewcommand{\arraystretch}{1.15}
\centering
\begin{threeparttable}
\scriptsize
\begin{tabularx}{\textwidth}{@{}l c l c X X@{}}
\toprule
\textbf{Sensor} & \textbf{DoF} & \textbf{Principle} & \textbf{Taxel number} & \textbf{Size} & \textbf{Range/Resolution} \\
\midrule
\textbf{This work}
& \textbf{3}
& \textbf{Magnetic (Hall)}
& \textbf{3}
& \textbf{2.50~$\times$~1.20~mm}
& \textbf{0--255~kPa; RMSE 2.13~kPa ($\sigma$=1.97)} \\

\cite{hou2024piezoresistive}
& 3
& Piezoresistive
& 1
& 4.5~$\times$~2.7~mm
& 1.2~N normal, 0.6~N shear; 0.228~mN (Z) resolution\\

\cite{usman2025_sensorized_grasper}
& 2
& Capacitive
& 1
& 6~$\times$~4~mm
& 0--5~N normal, 0--2~N shear; 0.0124~N resolution \\

\cite{li2024opticalwaveguide}
& 3
& Optical waveguide
& 1
& 2.6~$\times$~2.6~mm
& 0--3 N normal, shear; 0.089 N resolution \\

\cite{du2023design}
& 3
& Magnetic (Hall)
& 1
& $\sim$15~$\times$~15~mm
& 7.12~N normal; error 9.52\% \\

\cite{chua2023modular}
& 3
& Load cell
& 1
& 9.5~$\times$~8.5~mm
& $\pm$5~N axial, $\pm$3~N lateral; error $<$0.15~N ($<$11\%) \\

\cite{waters2022encouraging}
& 3
& Magnetic (Hall)
& 3
& 4~$\times$~9~mm
& 100\% (simulant), 77\% (liver) slip detection \\

\bottomrule
\end{tabularx}
\end{threeparttable}
\caption{\newtext{Comparison of jaw-mounted force sensors for surgical robot instruments.}}
\label{tab:sota-sensors}
\end{newtextblock}
\end{table*}

Contact interface sensors open the potential for a new approach in surgical grip control: sensing and predicting slip events.  
Incipient slip is the earliest stage of sliding when only part of the contact patch begins to move while the rest still adheres \cite{waters2020engineering}. 
Detecting slip in these regions allows the grasping force to be adjusted to prevent a full slip from occurring.
An initial approach presented a thermal-based sensor within the surgical jaw to detect movement of the tissue as slip began to occur \cite{burkhard2018slip}, but only offers a binary state of either slip or no slip. On-grasper optics have been used to detect slip in a surgical grasper \cite{lee2025reliable}, however \newtext{this approach} also only measures displacement.
Others have utilized distributed force sensing to monitor the progression of slip across a full grasping surface \cite{waters2020engineering, waters2022incipient}. The approach was successful for rudimentary grip control to prevent the slip \cite{waters2021utilising}; however, it utilized a large-scale model an order of magnitude too large for a real surgical instrument.
To \removed{the best of} our knowledge, localized tactile slip sensing directly on the jaw contact surface has not yet been demonstrated at this scale for RAMIS graspers.

In this letter, we present a replaceable on-jaw Hall-effect magnetic tactile sensor \removed{insert} for RAMIS graspers, capable of 3-DoF force and slip observability at the tissue contact interface. \removed{This sensor achieves high sensitivity considering the dimensional and magnetic constraints. From the dimensions of surgical grasper tool tips, and the goal of} Multiple taxels were needed to provide sufficient coverage of the grasping area \newtext{to detect slip and force across the grasping surface.} Given \newtext{tool-tip dimensions}\removed{the measurements of the surgical instrument}, PCB manufacturing constraints, Hall effect sensor size, \newtext{low-cost materials,} and \removed{maintaining a safe radius for magnets to minimize magnetic crosstalk between taxels} \newtext{the need to maintain a safe radius for magnets}, a design of three taxels was achieved\newtext{, resolving a mean pressure of 74.49 kPa.} \newtext{We validate the sensor on silicone phantoms and ex vivo tissue, showing local contact pressures and taxel-wise shear progression during incipient slip under clinically relevant loading, and further demonstrate live force profiles during human manipulation with \textit{da Vinci} teleoperation.}

\removed{The device integrates 2.50$\times$1.20$\times$2.50~mm individual magnetic taxels within a compliant surface, forming a fully sealed sensing module mounted on the grasper jaw.  
The taxel architecture supports estimation of both normal and shear forces during tissue retraction,} \removed{with 106~kPa resolved from the measured magnetic fields.}

\begin{figure}[h!]
    \centering
    \includegraphics[width=\linewidth]{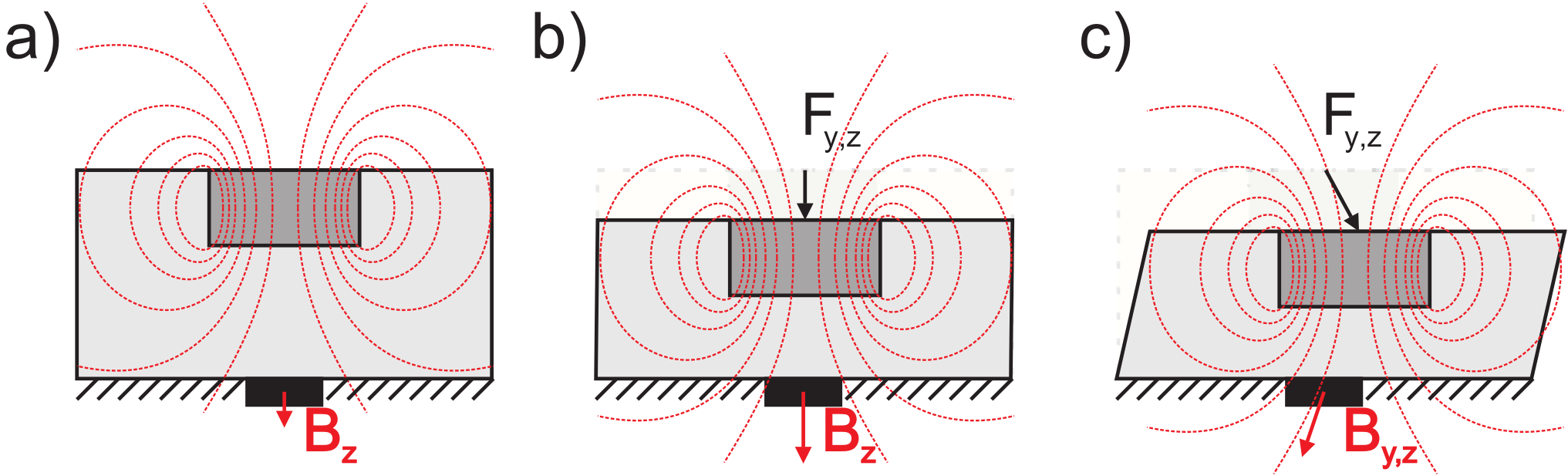}
    \caption{Sensing principle of Hall effect based sensors. a) Steady state measurement, b) Increase in contact pressure only displaces magnet vertically, giving a stronger field response ($B_z$); and c) A combined pressure and shear load displaces diagonally, changing the field response in both $B_y$ and $B_z$.} 
    \label{fig:sensor-working-principle}
\end{figure}

\section{Methods}

\subsection{Sensor design requirements}

To address problems of in-hand multi-axis force sensing on a miniaturized scale, we present a miniaturized force sensing array with three 3-DoF taxels, small enough to be affixed to the grasping jaw of a robotic surgical instrument. This device is small, lightweight, inexpensive, and enables sensing of grasp forces on the instrument jaws. The multiple taxels also enable directional slip sensing on the grasping face. By comparison, commercial sensors are bulky, costly, and lack the sensing density achieved in this work. 

To achieve force sensing on a surgical instrument, we target certain performance metrics: Detecting normal and shear forces within 0-2~N \cite{awad2024evaluation}, small sensor footprint, resolution of 0.1~N \cite{okamura2004methods}, \removed{and} a tunable design to fit the force range of the application \cite{wang2016design}\newtext{, and a low-cost scalable manufacturing process \cite{wang2016lowcost}}. \newtext{As tissue rarely spans the full jaw, sensing elements must cover the entire grasping surface, necessitating multiple taxels. More taxels enable local pressure measurement as well as cascading pressure changes over the interface, detecting slip. We maximize this to fit three taxels arranged along the principal axis of the jaw.}

\begin{figure*}[tb]
    \centering
    \includegraphics[width=\linewidth]{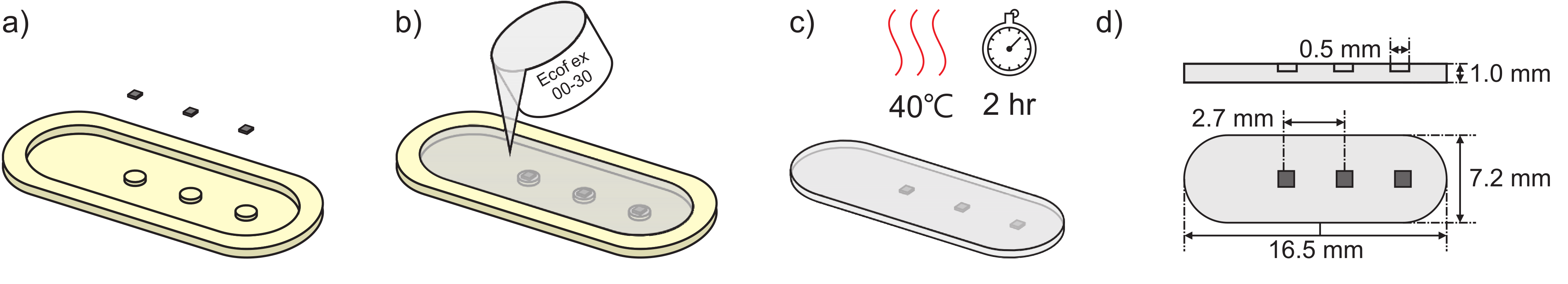}
    \caption{Design and manufacture of the tactile sensor array: Manufacturing process of the skin (a) place taxel magnets into mold, then (b) cast Ecoflex 00-30 elastomer overtop small magnets and cure, (c) remove cured elastomer with encased small magnets, leaving the (d) final profile.}
    \label{fig:elastomer-fabrication}
\end{figure*}

\subsection{Sensing principle}

The sensor operates using magnetic displacement sensing, where embedded magnets move relative to Hall sensors under applied force, producing measurable changes in magnetic flux density \cite{wang2016design}. The magnet islands are magnetized along their thickness and produce a field $B_{x,y,z}$. When the magnet is displaced by pressure and/or shear stress $F_{x,y,z}$, the magnetic islands will shift relative to their respective Hall sensors $d_{x,y,z}$ (Fig. \ref{fig:sensor-working-principle}). The displaced position of the magnetic island results in a change in the detected magnetic flux density of the Hall effect sensors beneath, which causes a voltage change in the Hall effect sensor. This voltage change is converted into magnetic field values onboard the sensor. 

Each taxel uses a pair of Hall sensors to provide a richer magnetic field measurement for force reconstruction. The proximal sensor pair are reserved for background stray field compensation, leaving three taxels in the current sensor design.

The force reference system of the sensor is shown in Fig. \ref{fig:sensor-overview}a, where the z-axis is normal to the sensor face, the y-axis is along the long axis of the sensor face (axis connecting the three taxels), and the x-axis is perpendicular to the y-axis on the sensor face.

\subsection{Sensor development}
In order to detect force and detect slip, we propose a Hall effect-based force sensor using magnetic displacement to measure applied normal and shear forces.

\begin{figure}[tb]
    \centering
    \includegraphics[width=0.9\linewidth]{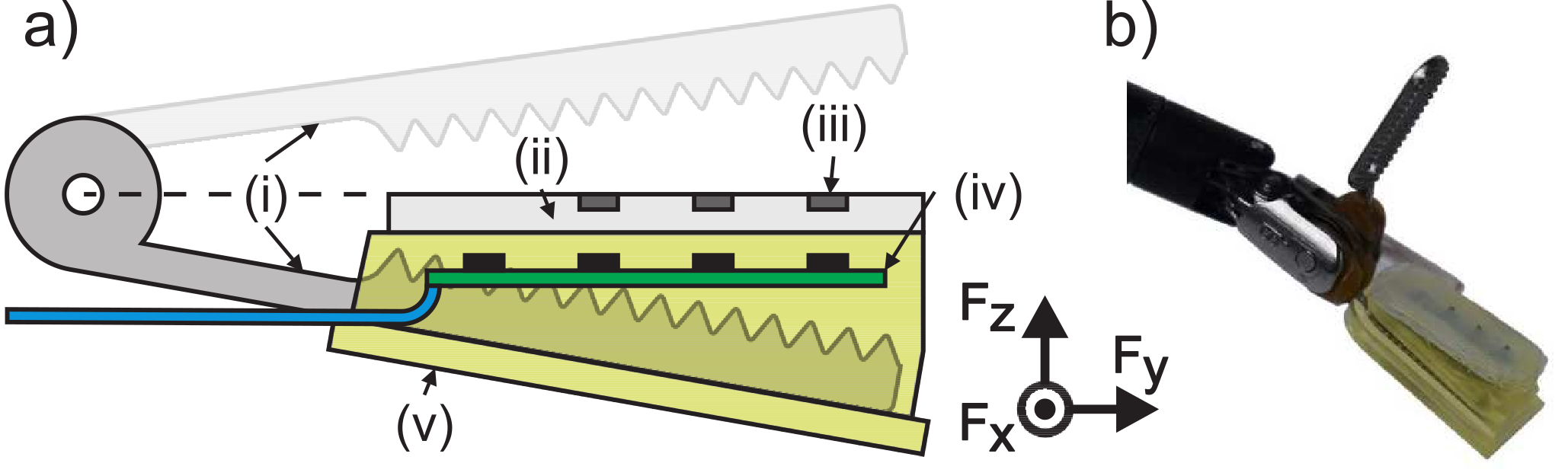}
    \caption{a) Schematic section view: (i) Original grasper jaw, (ii) elastomer layer, (iii) taxel magnet, (iv) PCB,  (v) 3D-printed mount; \removed{and (f) Sensor PCB indicating sensor nomenclature;} b) Sensor mounted in fenestrated forceps.}
    \label{fig:sensor-assembly}
\end{figure}

The assembled sensor has overall dimensions of 9.3$\times$6.5$\times$19.5~mm (Fig. \ref{fig:sensor-assembly}) and a mass of 2.06~g, with three individual taxels, each 2.50$\times$1.\removed{18}\newtext{20}$\times$2.\newtext{50}\removed{35}~mm. We use rigid NdFeB magnets for their high remanence, giving a measurable magnetic field signal at small scale.

To transduce the field, we use a custom PCB (2.5~mm$\times$12~mm$\times$0.5~mm) with eight 3-axis Hall effect sensors (AK09973D, Asahi Kasei Microdevices, Japan). 
The elastomeric sensing skin layer was fabricated using a 3D printed mold (HTL Resin, MicroArch 240, Boston Micro Fabrication, USA). \newtext{For each taxel, two} permanent magnets (0.5$\times$0.5$\times$0.2~mm; SuperMagnetMan, USA) are placed within the mold, held in position by a single larger permanent magnet on the exterior of the mold (Fig. \ref{fig:elastomer-fabrication}a). Ecoflex 00-30 elastomer (Smooth-On Inc., USA) is mixed (ARE-250 Thinkymixer, Thinky Corp., Japan), and vacuum degassed (5/01 Vacuum Caster, Renishaw, UK), before pouring into the mold and leveled (Fig. \ref{fig:elastomer-fabrication}b) then cured at 40\degree{}C. 
The cured elastomer is removed from the mold (Fig. \ref{fig:elastomer-fabrication}c), with a 0.5~mm elastomer thickness on one side and the magnets encased in the other (Fig. \ref{fig:elastomer-fabrication}d). The sensor skin is then affixed to a 3D-printed mounting housing the custom PCB (Fig. \ref{fig:sensor-assembly}a\removed{-c}\newtext{i-iv}). This assembly is mounted to the grasper jaw using a second 3D printed mount.

With the thickness of the elastomer layer underneath the magnets, the distance between the magnet and its respective sensor pair is 0.7 mm, a value determined by experimental evaluation of sensor sensitivity and desired responsiveness of the final tactile sensor. The maximum elastomer thickness of 1~mm was chosen to allow for adequate magnet displacement during contact ($\leq$50\%), while maintaining a small form factor. Furthermore, as the magnetic field decays cubically as a function of distance, the small magnets must be sufficiently close to the sensor \cite{wang2016design, petruska2013optimal}. Additionally, the contact surface has a 10\degree{} offset face from the grasper face (Fig. \ref{fig:sensor-assembly}a-b), to account for the thickness of the sensor, but reducing the maximum open jaw angle of 90\degree{} \cite{intuitive2007catalog}. As the sensor covers the existing teeth and fenestration of the tool jaw, the contact area is now the sensor face (107 mm$^2$). The sensor is connected to a microcontroller (Teensy 3.6, PJRC LLC., USA), which is connected to a PC for data recording over serial connection. The sensors are sampled at 500 Hz.

\subsection{Sensor calibration}
\begin{figure}
    \centering
    \includegraphics[width=0.95\linewidth]{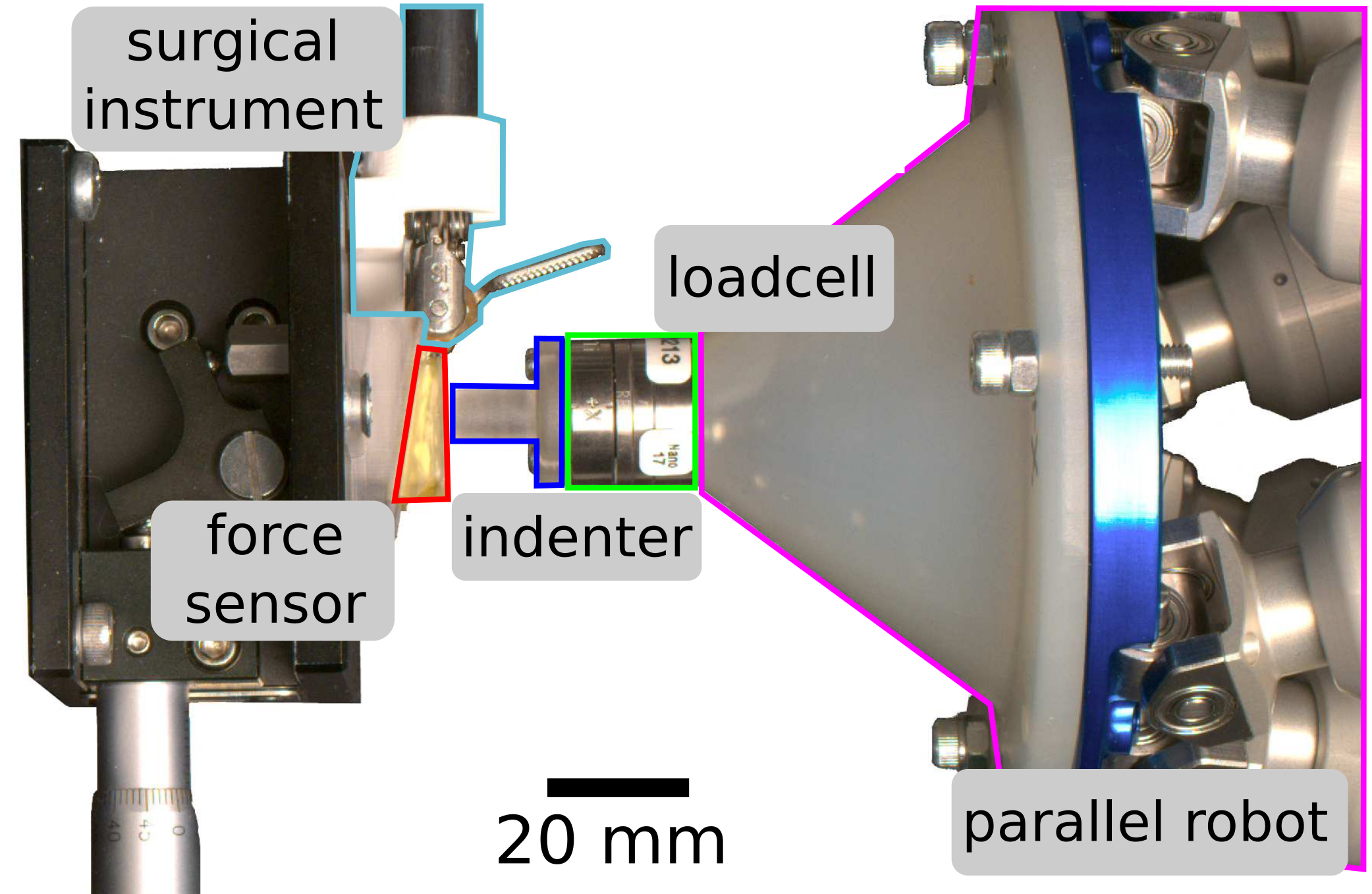}
    \caption{Calibration setup, top view.}
    \label{fig:calibration-side-top-view}
\end{figure}

A high-precision parallel robot platform (Solano, Symétrie, France) is used to apply precise deformations to the sensor elastomer surface. A \newtext{5 mm diameter 3D-printed circle indenter} on a 6-axis load cell (Nano 17, ATI, USA) is affixed to the end of the parallel robot to record the applied forces and torques to the sensor (Fig. \ref{fig:calibration-side-top-view}). Data acquisition is accomplished with a LabVIEW script (LabVIEW, National Instruments Corp., USA). Data is collated and synchronized with a MATLAB script (MATLAB, MathWorks Inc., USA). 

\begin{figure}
    \centering
    \includegraphics[width=0.95\linewidth]{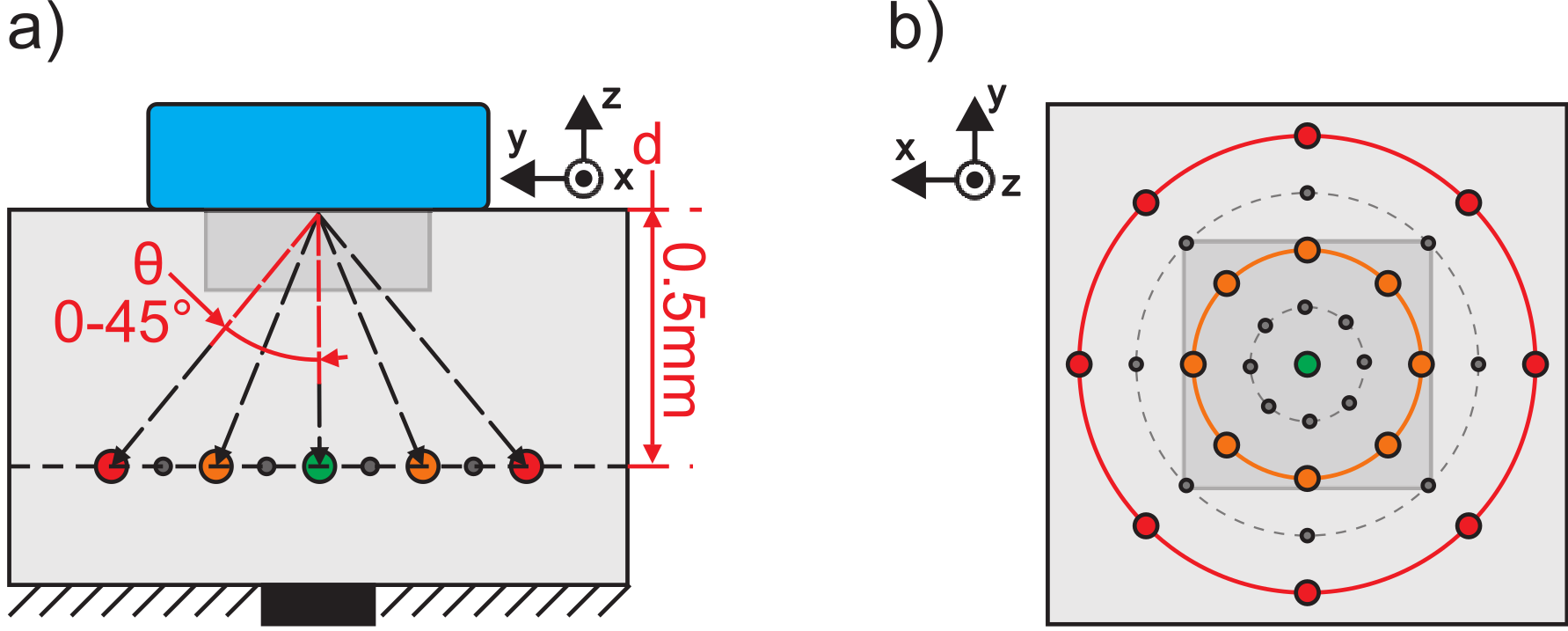}
    \caption{Calibration procedure at sensor surface (Not to Scale). The blue indenter is pressed into the surface to a depth ($d$) at an angle from the normal ($\theta$). Each value of theta is indented eight times around normal axis ($z$), forming several concentric cones. a) and b) show the cross sectional and planar endpoints.}
    \label{fig:calibration-traj}
\end{figure}

Calibration forces were applied along vectors spanning a cone centered on the surface normal. Indentation angles ranged from $0-45\degree{}$, producing combined normal and shear loads. Indentations reached a maximum depth of $d = 0.5$\,mm from the taxel surface (Fig.~\ref{fig:calibration-traj}a). All indentation vectors originated from the taxel normal axis and were distributed radially with a maximum incident angle $\theta = 45\degree$ (radius 0.5\,mm). Four concentric rings were defined around the normal axis, with linearly increasing radius and evenly spaced points, giving 21 total indentation locations (Fig.~\ref{fig:calibration-traj}d). 
The sensor was indented at a rate of 0.01~mm/s and sampled at 490 Hz. The calibration was downsampled and synchronized, yielding a rate of 150 Hz, totaling \textasciitilde\newtext{400,000}\removed{$403,855$}~datapoints per taxel for calibration. The indenter translated to each coordinate without rotational adjustment.

\begin{figure}[tb]
    \centering
    \includegraphics[width=\linewidth]{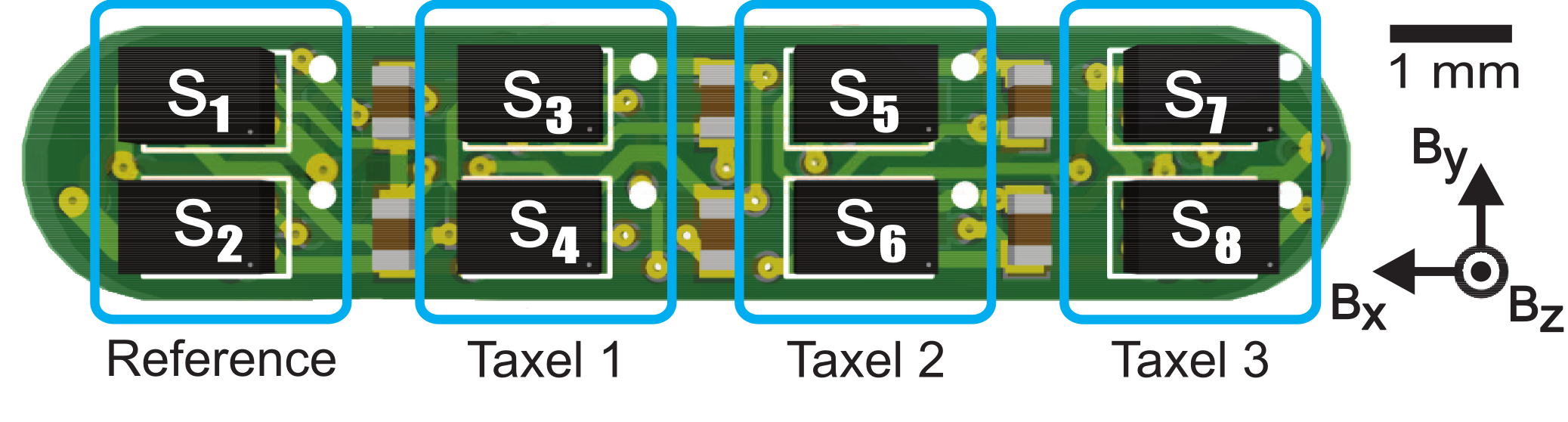}
    \caption{Force sensor PCB render with individual sensors and taxel pairs, and the sensor magnetic axes.}
    \label{fig:pcb}
\end{figure}

Before calibration, the load cell forces are filtered with a 3rd order Butterworth filter, a sampling frequency of 1~kHz, and a cutoff frequency of 10~Hz, to remove high frequency noise. Each taxel provides two sets of XYZ magnetic field measurements, denoted as $\mathbf{B}_{n}$ and $\mathbf{B}_{n+1}$ (Fig. \ref{fig:pcb}). 

To account for stray magnetic fields as the sensor changes orientation, the average measurement from the first pair of sensors ($s_{1}$ and $s_{2}$, most proximal along the sensor, Fig. \ref{fig:pcb} "Reference") is subtracted from the respective $\mathbf{B}$ measurement on each taxel. $\mathbf{B}$ values from the corresponding sensor pair of each taxel, along with the ground truth forces acquired during calibration indentations, are used to train a Feed-Forward Neural Network (NN) \cite{waters2022encouraging} using PyTorch \cite{paszke2019pytorch}, denoted as $\mathit{f}$ in (\ref{stray-field-eq}). A neural network regression model was used to capture the nonlinear relationship between magnetic field measurements and applied forces resulting from elastomer deformation and magnetic field coupling. The resulting calibration for taxel $i$ is $\mathbf{F_{i}}$ = ($\mathbf{Fx_{i}}$, $\mathbf{Fy_{i}}$, $\mathbf{Fz_{i}})$, where $i=[1,2,3]$. $\mathbf{F_{i}}$ are vectors containing timeseries force data resulting from the NN calibration, synced with the corresponding raw magnetic sensor values. The network consisted of a single hidden layer with 25 neurons using ReLU activation. Bayesian regularization was applied during training to reduce overfitting.

The network is trained to fit the tactile sensor magnetic data to the forces. \newtext{These forces are converted to pressure using the area of the calibration indenter (5~mm$\diameter$, A=19.635~mm$^2$), as the calibrations are performed with an area indenter rather than point forces.} Calibration results are shown in Table \ref{tab:calib-rmse}, with the lowest and highest RMSE in z being Taxel 3 and Taxel \newtext{2}\removed{1}, respectively. The calibrations are restricted by the resolution of the load cell (1.4~mN, 0.198~kPa in fx, fy, and fz).

\begin{table}[]
\centering
\begin{tabularx}{\linewidth}{@{} l *{3}{C} @{}}
\toprule
& \multicolumn{3}{c}{\textbf{Pressure Error (RMSE), kPa}}\\
\midrule
& \textbf{Taxel 1 \removed{(\textdownarrow)}} & \textbf{Taxel 2 \removed{(\textdownarrow)}} & \textbf{Taxel 3 \removed{(\textdownarrow)}} \\ 
\midrule
\textbf{fx RMSE kPa} & 1.4\removed{5}6 & \textbf{0.6\newtext{7}\removed{66}} & 0.74\removed{0} \\ 
\textbf{fy RMSE kPa} & \textbf{0.7\newtext{2}\removed{18}} & 0.77\removed{1} & 0.78\removed{1} \\ 
\textbf{fz RMSE kPa} & 4.8\removed{07} & 5.\newtext{4}\removed{382} & \textbf{3.\newtext{9}\removed{876}} \\
\bottomrule
\end{tabularx}
\caption{Calibration results by taxel.}
\label{tab:calib-rmse}
\end{table}

\begin{equation}\label{stray-field-eq}
\centering
\begin{array}{c}
    \mathbf{F}_{i}=f\left(\mathbf{B}_{n}-\mathbf{B}_{SF}, \mathbf{B}_{n+1}-\mathbf{B}_{SF} \right) \\
    \text{where:} \\
    \mathbf{B}_{SF} = \dfrac{\mathbf{B}_{s0} + \mathbf{B}_{s1}}{2}, i=[1,2,3], n=[2,4,6]  \\
    \end{array}
\end{equation}

To show the stray field compensation prevents sensor orientation from affecting the magnetic sensing, angular evaluation was performed. Starting in the zero home position (robot arm pointed downwards), the robot was driven from -40\degree{} to 40\degree{} on the first and second joints of the robot, equating to X and Y rotation of the sensor. The results show at most a 0.765~kPa deviation in force signal when no contact is applied (Table \ref{tab:stray-fields-mean-std}), within the range of angles.

\subsection{Characterization}\label{characterization}

\begin{figure}
    \centering
    \begin{subfigure}[t]{0.3\columnwidth}
        \centering
        \includegraphics[width=\linewidth]{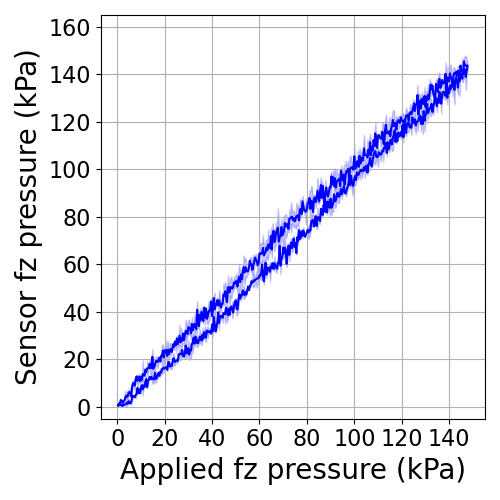}
    \end{subfigure}
    \begin{subfigure}[t]{0.3\columnwidth}
    \centering
        \includegraphics[width=\linewidth]{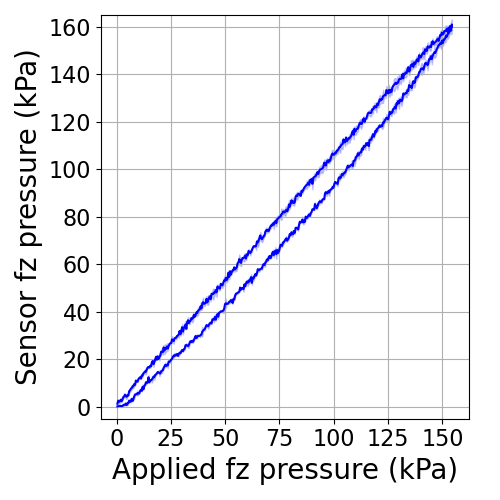}
    \end{subfigure}
    \begin{subfigure}[t]{0.3\columnwidth}
    \centering
        \includegraphics[width=\linewidth]{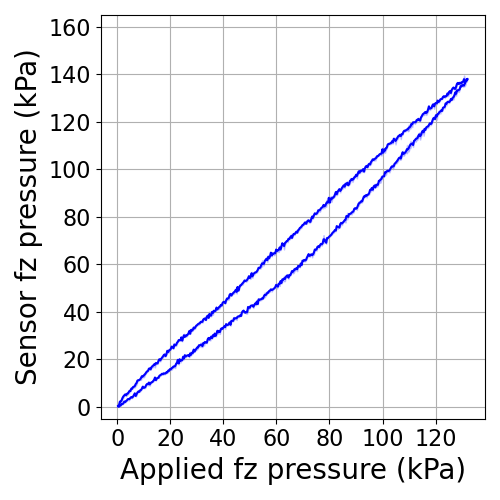}
    \end{subfigure}
    \caption*{Taxel 1 \hspace{40pt} Taxel 2 \hspace{40pt} Taxel 3}
    \vspace{1em} 
    \begin{subfigure}[t]{0.3\columnwidth}
    \centering
        \includegraphics[width=\linewidth]{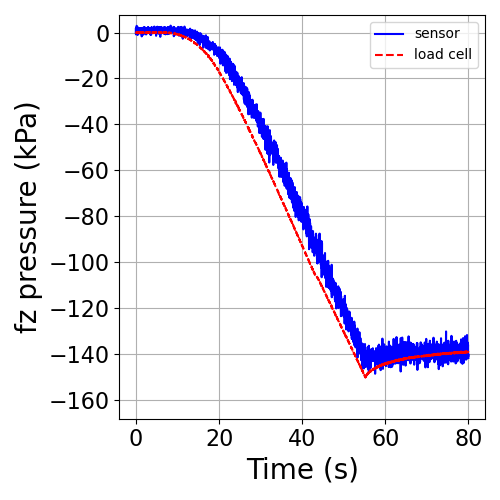}
    \end{subfigure}
    \begin{subfigure}[t]{0.3\columnwidth}
    \centering
        \includegraphics[width=\linewidth]{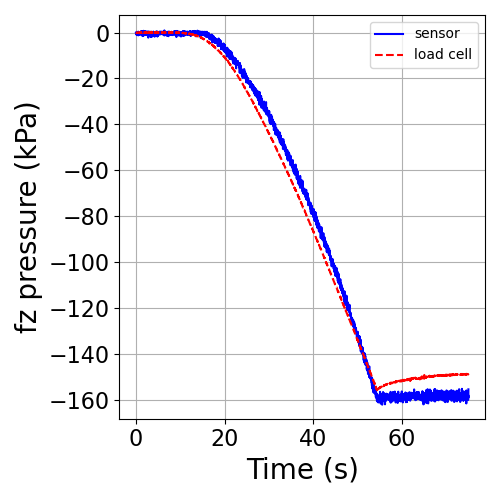}
    \end{subfigure}
    \begin{subfigure}[t]{0.3\columnwidth}
    \centering
        \includegraphics[width=\linewidth]{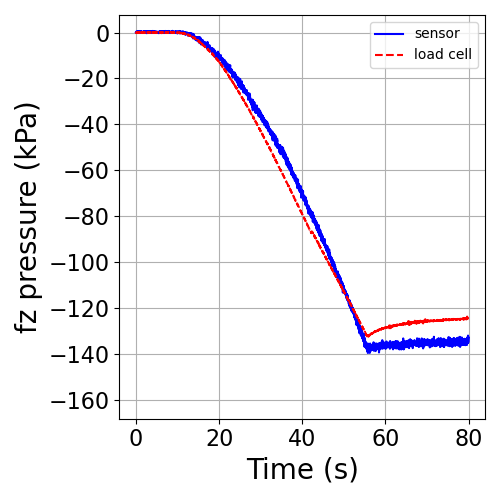}
    \end{subfigure}
    \caption{(top) hysteresis and (bottom) creep tests on each taxel.}
    \label{fig:sensor-characterisation}
\end{figure}

To assess the performance characteristics of the proposed sensor, a series of characterization tests were performed. Hysteresis tests in the sensor's Fz axis to a depth of 0.3~mm at 0.01~mm/s, n=10 tests were achieved (Fig. \ref{fig:sensor-characterisation} top), as well as creep tests indenting 0.3~mm and holding for 20~s were performed (Fig. \ref{fig:sensor-characterisation} bottom). The results of these are shown in Table \ref{tab:characterisation}.

\removed{Displacements up to 40\% were performed on the model, as beyond 50\% the contact mechanics do not converge well. These resulting magnetic field changes and their interactions between the taxel magnets are shown in Figure \ref{fig:mag-sim}.}

\begin{table}[]
\centering
    \begin{tabularx}{\linewidth}{@{} l *{4}{C} @{}}
    \toprule
    { \textbf{}} &
      { \textbf{Taxel 1}} &
      { \textbf{Taxel 2}} &
      { \textbf{Taxel 3}} \\ \hline
    { \textbf{Hysteresis (\%, {[}std{]})}} &
      { 14.75 {[}0.68{]}} &
      { 10.67 {[}0.39{]}} &
      { 13.55 {[}1.17{]}} \\ \hline
    { \textbf{Creep \newtext{(\% drop)}\removed{(RMSE, kPa)}} } &
      { \newtext{4.49} \removed{8.32} } &
      { \newtext{1.71} \removed{6.211}  } &
      { \newtext{3.81} \removed{6.405} } \\ \hline
    { \textbf{Creep (correlation)}} &
      { 0.996} &
      { 0.997} &
      { 0.995} \\ \bottomrule
    \end{tabularx}
    \caption{Characterization results.}
    \label{tab:characterisation}
\end{table}

Given the tight tolerances required to fit in the small grasper jaw, submillimeter magnets were selected. To validate the magnetic behavior of the taxel magnets, a FEM simulation (COMSOL Multiphysics, COMSOL Inc., Burlington, MA, USA) was developed. \removed{(Fig. \ref{fig:mag-sim})} The dimensional and magnetic parameters of the magnets (0.5$\times$0.5$\times$\newtext{(2x 0.2~mm)}, neodymium, N50) were input to the simulation, which showed minimal interaction between their fields even when displaced \newtext{(as Taxel 2 was displaced 0.5~mm Z-, Taxel 1 had maximum magnetic field changes in Bx, By, Bz of 14\%, 7\%, 13\% while Taxel 3: 26\%, 9\%, 10\%, respectively)}. \newtext{In only Y-axis displacement, $\Delta B_{y}$ of the adjacent taxels was still at most 26\%.} 

\newtext{While increasing the magnet size (1x, 1.5x, 2x) shows increased sensitivity in the Y and Z-axes (1x: 4.15~mT, 1.5x: 13.42~mT, 2x: 15.55~mT), respective crosstalk generally increases (1x: y-15.4\%, z-18.3\%; 1.5x: y-13.9\% z-12.8\%; 2x: y-15.3\% z-17.3\%).

This indicates that the selected magnet is a suitable size to achieve required sensitivity, and well within these upper bounds of magnetic crosstalk}. The \removed{ir} \newtext{taxel magnets are arranged in a} North-South-North orientation\newtext{, which} encourages the \newtext{magnetic} fields to "flatten" \newtext{or be pulled closer to the magnet surface plane}, restricting the flux \newtext{from reaching adjacent sensors or magnets}\removed{(Fig. \ref{fig:mag-sim})}.

\begin{table}[]
    \centering
    \begin{tabularx}{\linewidth}{@{} l *{3}{C} @{}}
    \toprule
    & \multicolumn{3}{c}{\textbf{Pressure Mean (Std), kPa}}\\
    \midrule
    & \textbf{Taxel 1} & \textbf{Taxel 2} & \textbf{Taxel 3} \\ 
    \midrule
    \textbf{fx kPa} & 0.517 (0.179)   &  1.137 (0.122) &  -3.905 (0.279) \\ 
    \textbf{fy kPa} & -2.634 (0.329)  & -1.296 (0.503) &   5.611 (0.318) \\ 
    \textbf{fz kPa} & -12.281 (0.650) & -6.427 (0.879) & -15.020 (2.745) \\
    \bottomrule
    \end{tabularx}
    \caption{Mean and standard deviation during angular displacement [-40\degree{}, +40\degree{}].}
    \label{tab:stray-fields-mean-std}
\end{table}

\newtext{Magnetic crosstalk is observed with each taxel indented in a 5-point trajectory (X$\pm$0.5~mm, Y$\pm$0.5~mm, Z+0.5~mm). For the other two taxels, the maximum deviation from the magnetic field value at rest was computed, shown in Table \ref{tab:crosstalk}. Taxel 1 experiences the largest crosstalk, at 8.1\% and 9.9\% from Taxel 2 and 3, respectively.}

\renewcommand{\arraystretch}{0.85}
\begin{table}[]
    \centering
    \begin{tabularx}{\linewidth}{@{} l *{3}{C} @{}}
    \toprule
     & \textbf{Taxel 1} & \textbf{Taxel 2} & \textbf{Taxel 3} \\
    \midrule
    Taxel 1 & - & 8.1 & 9.9 \\
    Taxel 2 & 5.6 & - & 5.7 \\
    Taxel 3 & 2.2 & 3.9 & - \\
    \bottomrule
    \end{tabularx}
    \caption{Crosstalk (\%) between taxels, Z-axis.}
    \label{tab:crosstalk}
\end{table}

\newtext{To understand the effect of different size indenters on taxel calibration and validation, five different indenters (2, 3, 4, 5, 20~mm) were used to calibrate and test the taxel. Tests of indenter calibrations on the same sizes yielded low RMSE (under 1.5~kPa), however mismatched tests found large errors. Smaller indenter calibration models under-predict on larger test data, and vice versa (around $\pm$18~kPa at the extremes). Direction of mismatch flips the sign of the bias, meaning negative for models calibrated on larger indenter than the test, positive for models calibrated on smaller indenter than the test.}

\section{Materials}
\subsection{Robotic Evaluation}
\begin{figure}
    \centering
    \includegraphics[width=0.7\linewidth]{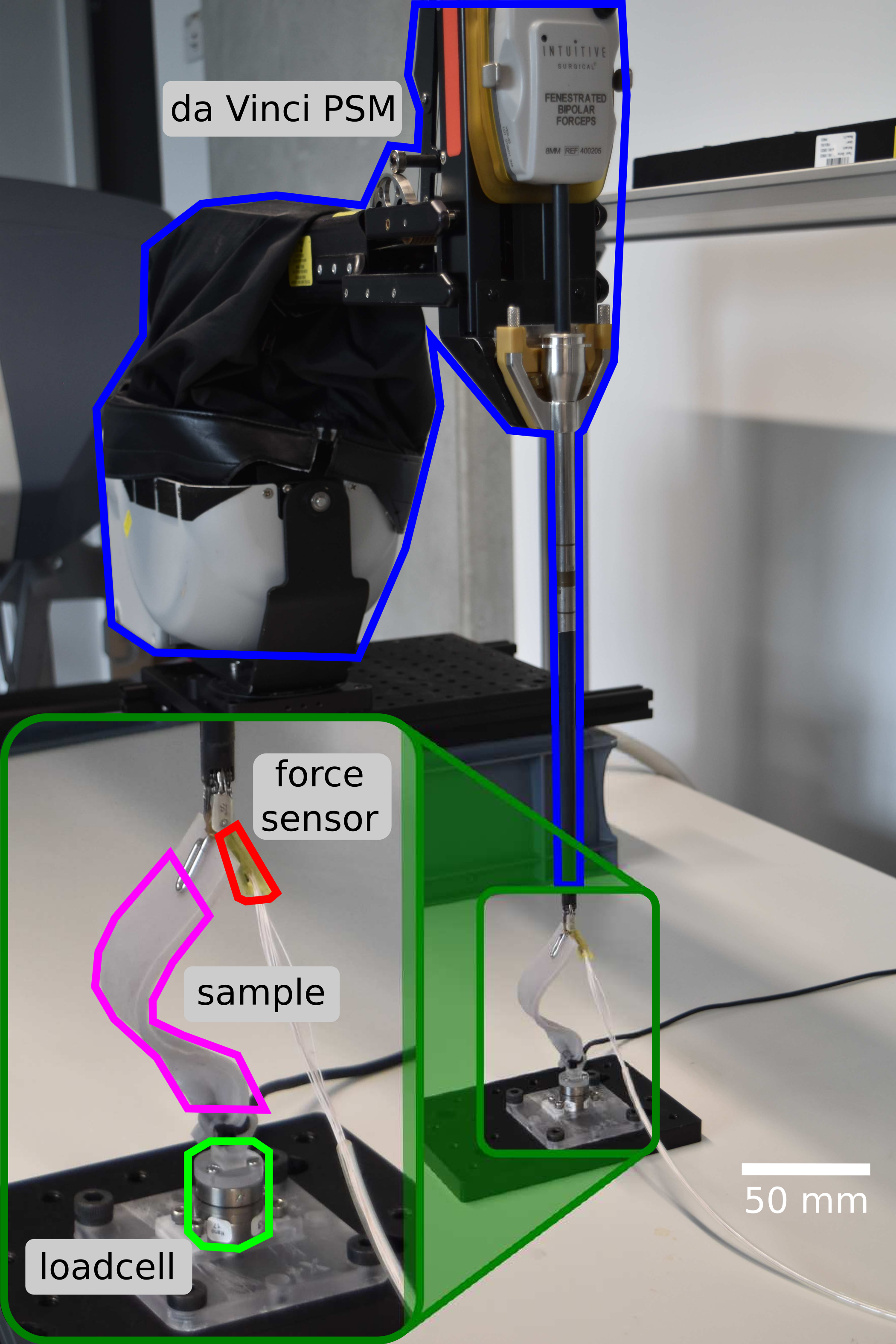}
    \caption{Experimental setup with \textit{da Vinci} fenestrated bipolar forceps instrument.}
    \label{fig:exp-setup}
\end{figure}

To evaluate the sensor in-situ, we utilize a da Vinci Research Kit \cite{kazanzides2014opensource}. The sensor was affixed to one jaw of a pair of fenestrated bipolar forceps (400205, Intuitive Surgical, USA) mounted on the \textit{da Vinci} Patient Side Manipulator (PSM).
This instrument then grasps and pulls on the respective sample, the other end of which is affixed to a bench-mounted 6-DoF load cell (Nano17, ATI, USA). The PSM's instrument principal axis (i.e. the tool shaft) was aligned along the Z-axis of the load cell, allowing for direct measurement of the pull force along the y-axis of the sensor (Fig. \ref{fig:exp-setup}). \newtext{The load cell is not a direct comparison to the sensor values, but rather a reference quantification of the bulk material response in the grasper, as measured by the proposed sensor \cite{waters2022incipient}.}

To enhance the sensor signal, a 3D-printed wedge toothed jaw part was made and affixed to the opposite grasper jaw. Removing the fenestration while maintaining the same toothed design allowed for all areas of the sensing surface to monitor pressure rather than focused pressure around the perimeter \cite{guo2022friction}\newtext{. This} \removed{and} serves as a surface against the grasped material to improve sensor readings, as opposed to the typical fenestrated design. 

The same experimental setup as in Fig. \ref{fig:exp-setup} was used to test slip. The silicone simulant and tissue samples were grasped and retracted until a slip condition was observed, as pull force dominates frictional force. Each silicone sample had lubricant applied to emulate the soft wet conditions of living tissue, improving slip. Video of an exemplar tissue slip experiment can be seen in supplementary materials (video S1). 

In this work we focus on low-frequency slip, which is characteristic of soft tissue interactions, as high-frequency slip is relevant more in rigid-body interactions. For this we can use the metric of the Coefficient of Friction for each taxel's $F_y$ direction, as $F_{{y}_{CoF}} = F_y/F_z$, where $F_y$ is the y-axis shear force on the taxel, and $F_z$ is the normal force on the taxel, along the z-axis \cite{waters2020engineering}. If this ratio makes a sharp change, that is indicative of a slip event; where $F_y$ force dominates, the $z$ force is insufficient to maintain a grasp.

\subsection{Experimental Samples}

\subsubsection{Silicone tissue simulant}
Our initial tests were performed on 100~mm$\times$20~mm$\times$3~mm pieces of Ecoflex 00-20 with 2 layers of fabric as a strain limiting layer to mimic organic tissue (from \cite{waters2020engineering}). The silicone sample was affixed to the load cell in a vertical orientation, with the upper end being grasped and pulled by the robotic surgical instrument (Fig. \ref{fig:exp-setup}).

\subsubsection{Ex vivo tissue}
To supplement the silicone simulant experiments, grasping was assessed on porcine small intestine (ileum) tissue. Tissue samples were procured from the food chain and cut into roughly 40$\times$5$\times$2~mm pieces for testing, and were stored in Phosphate-Buffered Saline to maintain hydration.

\section{Results}

\subsection{Silicone}
\begin{figure}
    \centering
    \begin{subfigure}[t]{0.95\linewidth}
        \centering
        \includegraphics[width=\linewidth]{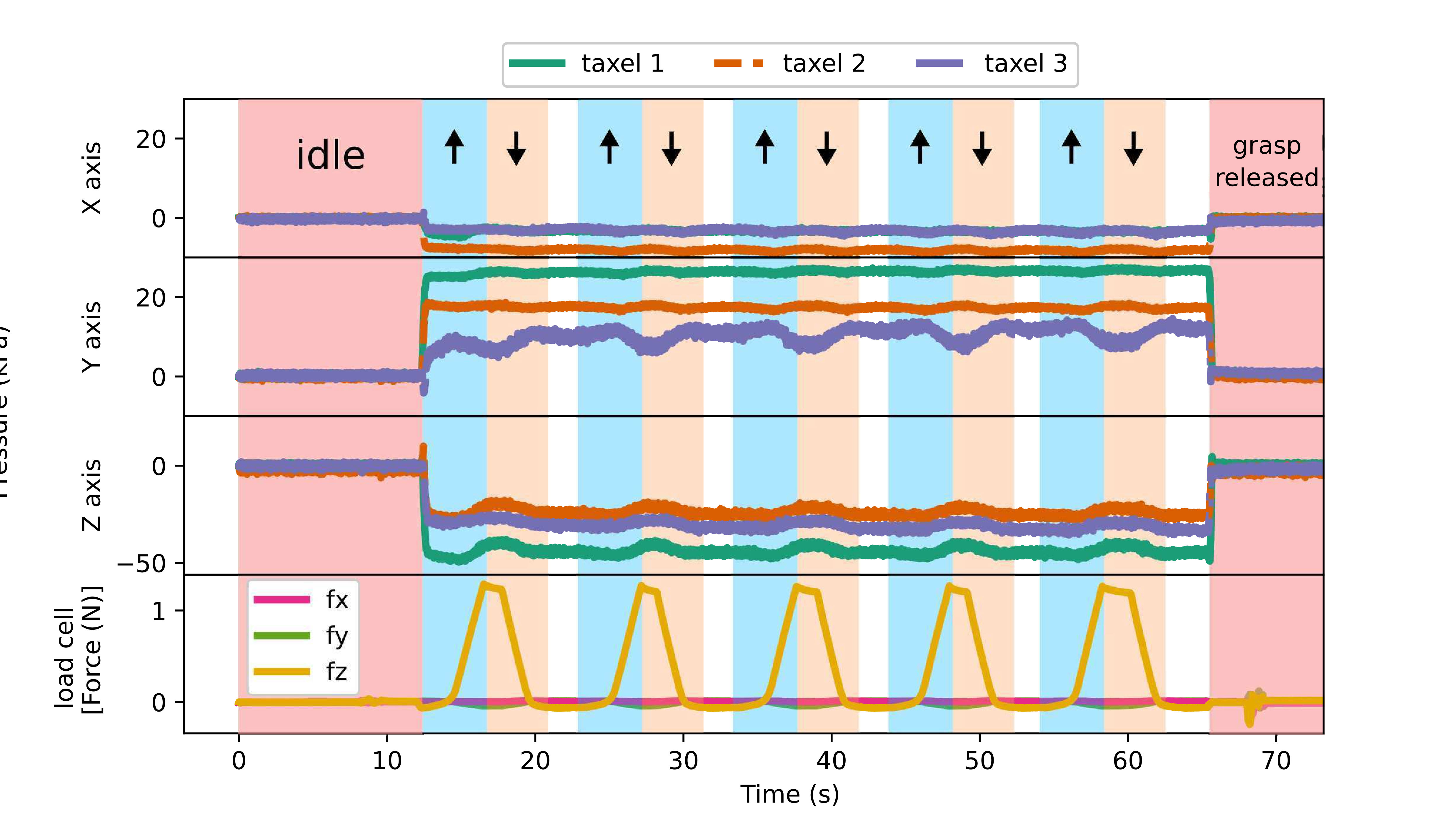}
        \caption{Cyclic retraction on silicone simulant, predicted forces by taxel for each axis with ground truth load cell.}
        \label{fig:retr-exp-2-pred-fxyz-three-taxels}
    \end{subfigure}

    \begin{subfigure}[t]{0.95\linewidth}
        \centering
        \includegraphics[width=\linewidth]{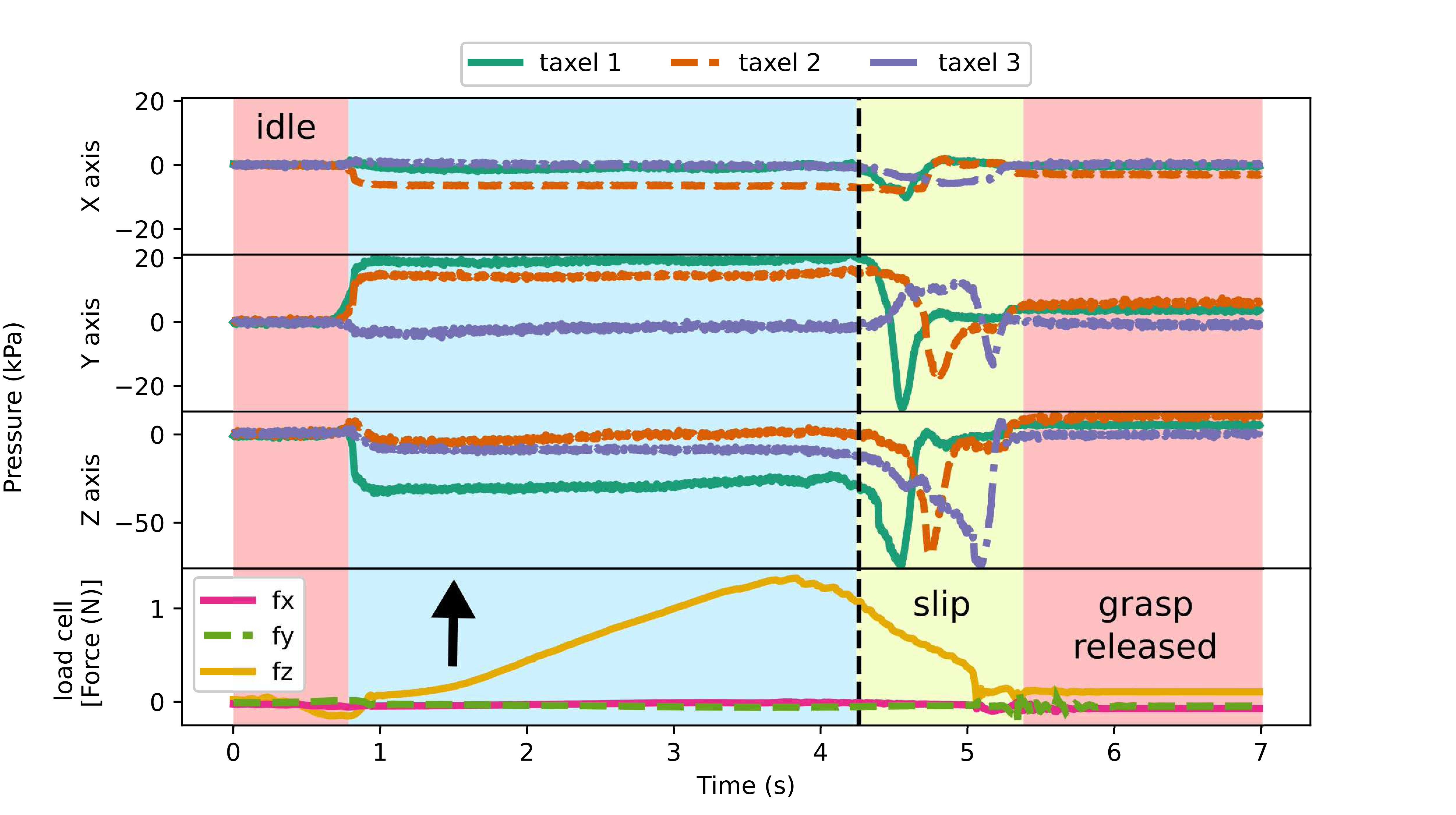}
        \caption{Slip experiment on silicone tissue simulant, 0.1~mm/s. Dashed vertical line denoting slip event.}
        \label{fig:slip-experiment-2}
    \end{subfigure}
    \caption{Retraction and slip experiments on silicone tissue simulant.}
    \label{fig:combined-experiments}
\end{figure}

Retraction of the silicone simulant \removed{(n=10)} shows clear applied forces in X, Y, and Z, as well as the cyclic tugging force as the material is pulled (Fig. \ref{fig:retr-exp-2-pred-fxyz-three-taxels}). \newtext{Peak pressures were 27.5 kPa in Y and 91.2 kPa in Z, and in each experiment were highly repeatable across 5 pulls (Std: < 0.28~kPa on Y and Z).} \removed{Peak positive forces reached 1.96~kPa in X, 27.4~kPa in Y, and 4.52~kPa in Z, with very low standard deviations between retractions (< 0.2~kPa). Minor pressures in X (\textasciitilde{}-8~kPa) are likely due to variations in alignment of the sensors and the load cell.}

Slip experiments showed force change and slip at sequential taxels, with slip onset occurring at \textasciitilde{}14.25~kPa shear force in the Y-axis on Taxel 1, $t\approx 4.2 s$. The slip propagated distally along the sensor (Fig. \ref{fig:slip-experiment-2}), with peak Y-axis shear forces of \textasciitilde{}-31.59~kPa, \textasciitilde{}-27.41~kPa, and \textasciitilde{}-8.14~kPa on Taxels 1, 2, and 3 \newtext{respectively}\removed{consecutively}.

\subsection{Tissue}
\begin{figure}[t]
    \centering
    \begin{subfigure}[t]{0.95\linewidth}
        \centering
        \includegraphics[width=\linewidth]{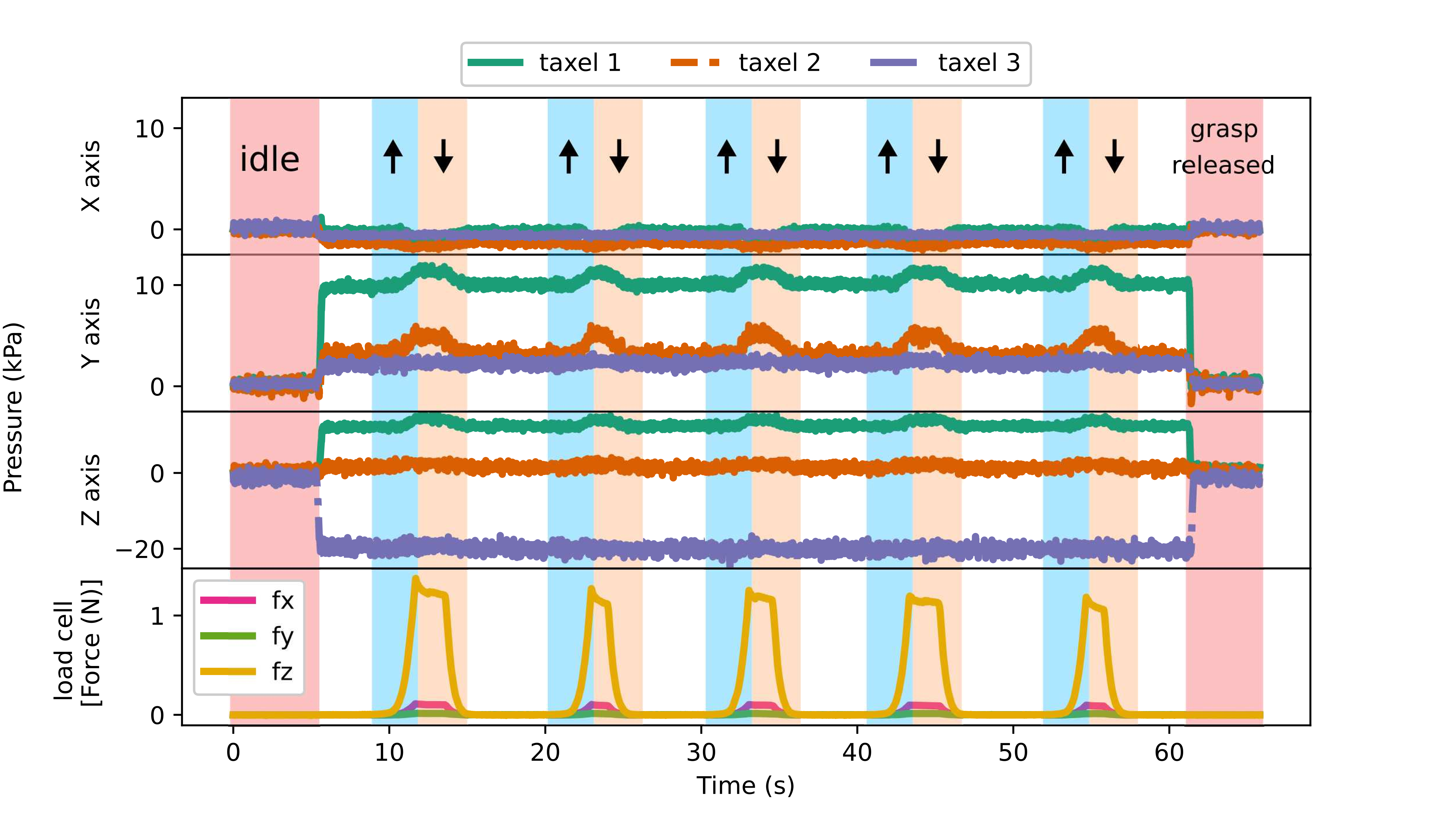}
        \caption{Retraction experiment on porcine tissue, 0.1~mm/s over 15 mm.}
        \label{fig:retr-experiment-tissue-9-20251002}
    \end{subfigure}
    \begin{subfigure}[t]{0.95\linewidth}
        \centering
        \includegraphics[width=\linewidth]{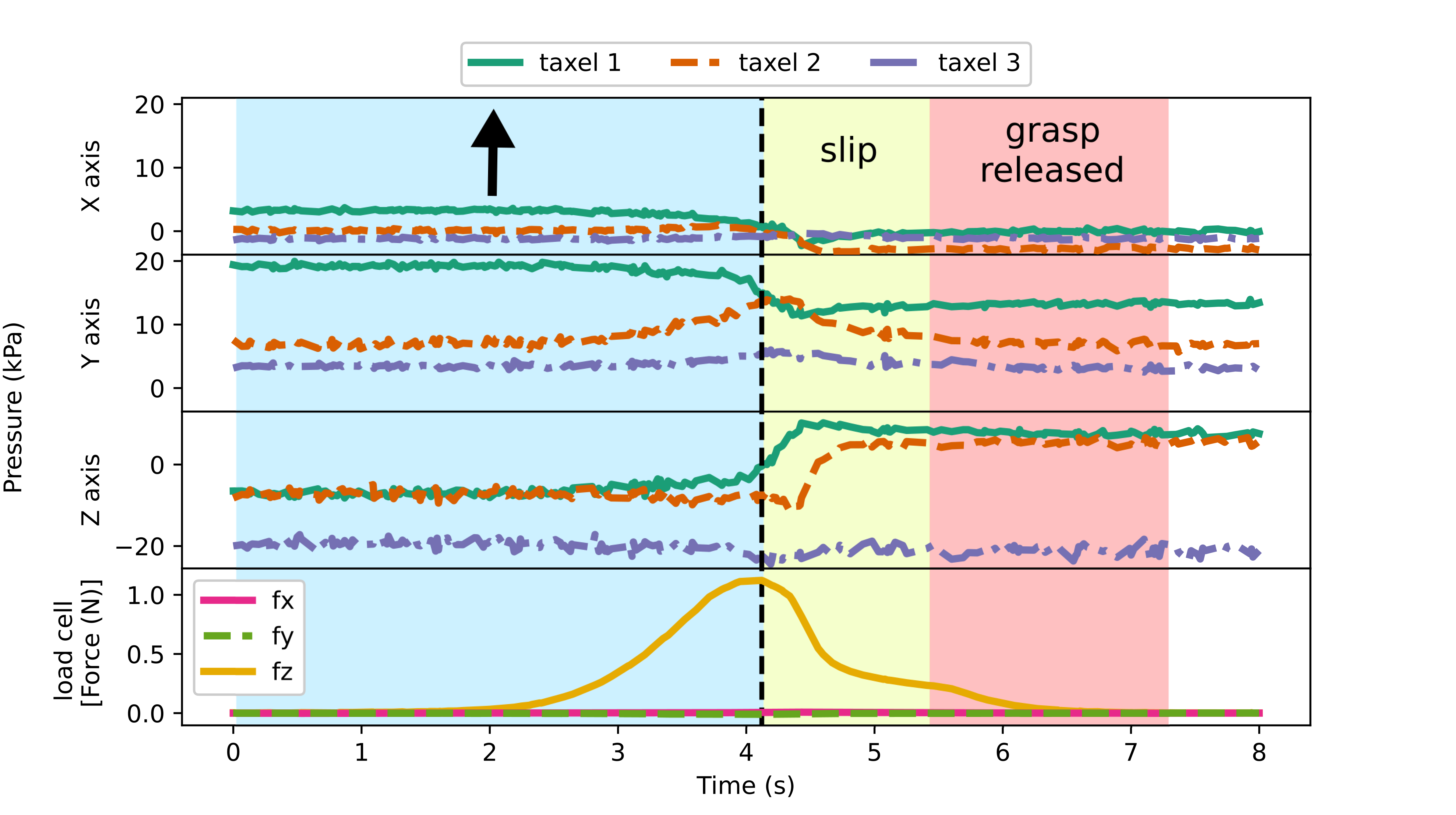}
        \caption{Porcine small intestine slip, 0.1~mm/s. Dashed vertical line denote slip event.}
        \label{fig:tissue-slip-exp}
    \end{subfigure}
    \caption{Retraction and slip experiments on porcine tissue.}
    \label{fig:combined-tissue-experiments}
\end{figure}

Retraction of the tissue yielded similar results as in silicone, seeing the peaks in pull force in all directions but primarily in Y, then Z. These are likely due to the compressive jawed grasp of the tool being greater than the frictional forces of the tissue on itself, allowing it to shift somewhat when pulled; the sensors are able to detect this change, however (Fig. \ref{fig:retr-experiment-tissue-9-20251002}).

Tissue retraction shows lower peak pressures than silicone, with maxima values of 11.9~kPa [Std: 0.038 kPa] in the Y-axis and 15.3~kPa [Std: 0.081 kPa] in the Z-axis. Tissue maximum pressures are around 50\% lower than in silicone, reflecting the impact of lower stiffness and reduction in shear force transfer of soft, wet tissue during retraction.

In Fig. \ref{fig:retr-experiment-tissue-9-20251002}), bunching of the tissue in the grasper jaw and imperfect alignment with the taxel magnets could cause magnet tilt, resulting in an apparent increase in force in Taxel 1 Y axis.

We contrasted tissue retraction at 10 mm and 15 mm to observe an increased pull force, which showed a \textasciitilde2.5x increase in max pull force (mean [std] : 29.592~kPa [3.974 kPa] versus 74.49~kPa [11.939 kPa]). This shows the sensor can maintain a solid grasp on tissue while still providing shear force information.

In tissue slip experiments (Fig. \ref{fig:tissue-slip-exp}) we see a similar hyperelastic response, as the tissue slips in the \removed{instrument} jaws. The tissue does not leave the grasper, however, and as the tool returns to the home position the load cell Z force returns to zero. 

Tissue slip showed more irregular, gradual forces than silicone, indicating the heterogeneous biological material responds differently to load. The repeated observation of gradual, irregular force responses across tissue trials suggests that the sensor is responsive to biologically relevant loading conditions, in this case ex vivo porcine intestine.

Variation analysis of the slip experiments (n=11 \newtext{samples}) shows the asymmetrical performance of the taxels across the sensor face. While Taxel 1 was most consistent\newtext{, with a coefficient of variation (CV) of} \removed{(CV}<0.64\removed{)}, Taxel 2 had high variability (CV>1.6). 
This is understandable given the grasping geometry is a scissor grasp, resulting in higher forces towards the hinge point. These higher forces could result in more stable grasping, therefore less variation during manipulation. Y-axis forces were the most reproducible, and the Z-axis forces were the most variable across the taxels, pointing to variable force distribution during slip events.

\subsection{Freeform tissue grasping}
\begin{figure}
    \centering
    \includegraphics[width=0.95\linewidth]{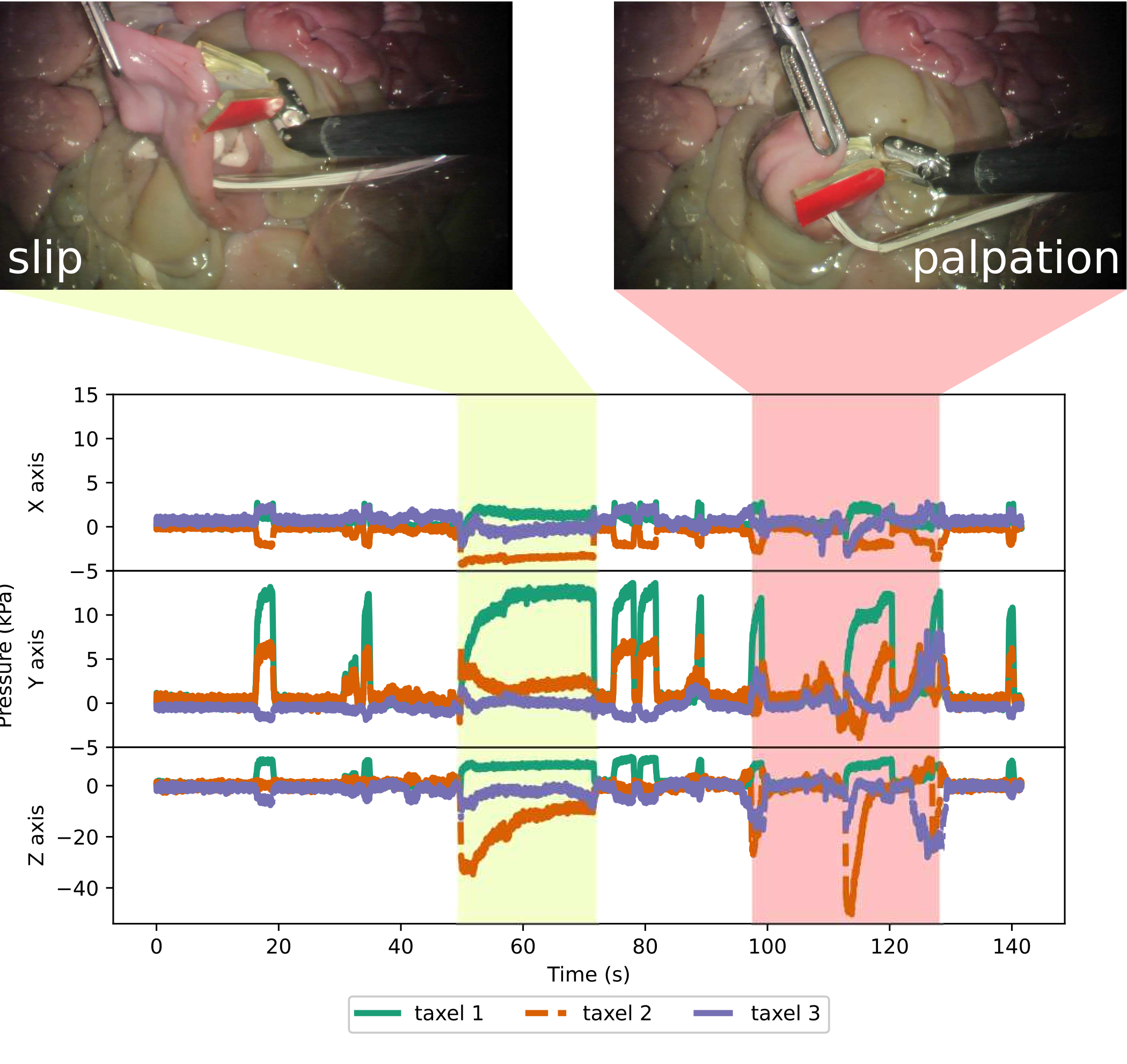}
    \caption{Freeform tissue grasping experiment, showing tissue slip during manipulation and repeated tissue palpation.}
    \label{fig:freeform-exp}
\end{figure}

To demonstrate the sensor in more realistic scenarios, we devised manipulation experiments with ex vivo porcine gastrointestinal tract. The first manipulation experiment consisted of grasping and handoff of the tissue between the sensorized robotic instrument and a standard laparoscopic tool. In the second experiment, the laparoscopic tool held the tissue while the sensorized instrument grasped and palpated the tissue. These experiments can be found in the supplementary materials (videos S2 and S3), and in Figure \ref{fig:freeform-exp}. These demonstrations show that the sensor can generate stable multi-taxel force traces during unconstrained handling tasks beyond the one-dimensional benchtop retraction experiments.

\section{Discussion}
A key outcome of this work is demonstrating that magnetic tactile sensing can be miniaturized to fit within the physical constraints of a surgical grasper jaw. The sensors presented achieved at least \removed{3}\newtext{6}\% resolution of the max pressure\removed{(comparing highest calibration RMSE to max pressure detected: 139~kPa)}\newtext{, detecting retraction max pressures of 74.49 kPa}. The sensors are also capable of detecting forces within the targeted range of 0-255~kPa, typical in RAMIS procedures \cite{awad2024evaluation}. \newtext{Calibration errors in each axis also align with 0.1 N sensor resolution (Table \ref{tab:calib-rmse}).} Varying the magnets used, \removed{the} distances from the Hall sensors, and the elastomer used can tune the sensitivity of the sensors for different applications and force ranges. While the sensor enables grasp-site force sensing, it also adds to the surgical instrument footprint. 

The flat, smooth elastomer face of the sensor permits more slip than the unmodified fenestrated toothed jaw, \newtext{resulting in grasps} \removed{and therefore the achieved grasps are} not as strong and robust as before modification. The consequence of elastomer-based sensor design is the interface of hyperelastic tissue with hyperelastic elastomer, making their behavior more difficult to ascertain. Future iterations will address sterilization requirements and sealing to prevent fluid ingress.

The variation of tissue traits (variable thickness, hydration, elasticity, etc.) leads to variations in achieved grasp. More factors affect the quality of grasping and slip data, such as tissue type, contact area with the sensor, and pull angle/plane. \removed{Freeform manipulation experiments made use of external tools to position tissue, as the sensor limited the maximum grasp angle the robotic instrument could reach.} The retraction and slip experiments focused on the principal axis where slip would occur, \removed{however this is} a simplification from clinical use. 

The $y_{CoF}$ based slip detection metric displayed slip in experiments on silicone and tissue. In silicone, the $y_{CoF}$ peaks aligned well with the ground truth fz forces, showing distinct slip over consecutive taxels. However in tissue, the $y_{CoF}$ values varied greatly, indicating non-uniform and variable slip interactions. This may arise from inconsistent contact, variable tissue properties across the sample, stick-slip friction phenomena. These variations highlight the complexity of tissue mechanics and motivate further investigation into sensor behavior across different tissue types.

\section{Conclusions}

In this work we present a miniaturized Hall-effect tactile sensor integrated directly into the grasping surface of a robotic surgical instrument, demonstrating that multi-axis tactile sensing can be implemented at millimeter scale within surgical instrument constraints. The sensor was calibrated with ground truth force data and applied to a surgical robotic grasper. Experimental \newtext{demonstration} \removed{validation} on ex vivo porcine tissue showed the grasping force sensing capabilities of the sensor, as well as detection of slip during grasping.

The \newtext{versatile} sensing architecture \removed{may}\newtext{will} allow \newtext{for customizable taxel configurations, and is scalable for high volume production as a disposable add-on sensor.} \removed{future designs with customized taxel configurations. Future designs will feature taxels spanning width-wise, so as to measure incipient slip more readily at the sides of the contact region.} Further work would also explore the effects of non-normal angle indentations to the tactile sensors, achieving more detailed calibrations. 

In this iteration, one grasping surface is sensorized, and this modification likely altered the performance of the original grasping capabilities of the instrument. Further development would evaluate the modified tool grasp to minimize performance interference from the added sensor. Sterilization is a concern for surgical devices, and such considerations will influence future material choices. Development towards clinical viability will explore biocompatible materials for the elastomer and sensor enclosure, or use a minimal single-use design. More intricate magnetic profiles would provide distinct magnetic regions for the sensors, enabling greater resolution and sensitivity.

\renewcommand\bibname{References}
\printbibliography

\end{document}